\documentclass[11pt,a4paper]{article}
\usepackage[hyperref]{emnlp2018}
\usepackage{times}
\usepackage{latexsym}

\usepackage{url}
\usepackage{tabularx}

\aclfinalcopy 

\def\Snospace~{\S{}}  

\usepackage{amsmath, amssymb}
\usepackage{booktabs}
\usepackage{dirtytalk}  
\usepackage[inline]{enumitem}
\usepackage{graphicx}
\usepackage{microtype}
\usepackage{subcaption}
\usepackage{siunitx} 
\DeclareSIUnit{\nothing}{\relax}
\DeclareSIUnit{\bleu}{\textrm{BLEU}}

\usepackage{multirow}

\usepackage{tikz}
\usetikzlibrary{shapes.geometric}
\usetikzlibrary{positioning}
\usetikzlibrary{patterns}
\usepackage{pgfplots}
\pgfplotsset{compat=1.14}

\newcommand{\pair}[2]{#1--#2}
\newcommand{\toEn}[1]{\pair{#1}{En}}

\usepackage{letltxmacro}
\LetLtxMacro{\oldhref}{\href}
\renewcommand{\href}[2]{\oldhref{#1}{\texttt{#2}}}

\usepackage{thanks-nostar}

 \title{Freezing Subnetworks to Analyze Domain Adaptation \\ in Neural Machine Translation }

\author{}

\author{Brian Thompson$^\dagger~~$ Huda Khayrallah$^\dagger~~$ Antonios Anastasopoulos$^\ddagger$ \\ \bf{ Arya D. McCarthy$^\dagger~~$ Kevin Duh$^\dagger~~$ Rebecca Marvin$^\dagger~~$ Paul McNamee$^\dagger$} \\ \bf{Jeremy Gwinnup$^\circ~~$ Tim Anderson$^\circ~~$ \textnormal{and} Philipp Koehn$^\dagger$}\\
$^\dagger$Johns Hopkins University,  $^\ddagger$University of Notre Dame, $^\circ$Air Force Research Laboratory \\
  {\tt \{brian.thompson, huda, arya, becky, mcnamee, phi\}@jhu.edu,} 
  \\  {\tt aanastas@nd.edu,    kevinduh@cs.jhu.edu,} \\ {\tt \{jeremy.gwinnup.1, timothy.anderson.20\}@us.af.mil} \\
}
\date{}

\begin{document}

\maketitle
\begin{abstract}

To better understand the effectiveness of continued training, 
we analyze the major components of a neural machine translation system
(the encoder, decoder, and each embedding space) and consider each component's contribution to, and capacity for, domain adaptation. 
We find that freezing any single component 
during continued training
has minimal impact on performance, 
and that performance is surprisingly good 
when a single component is adapted while
holding the rest of the model fixed. 
We also find that continued training does not move the model very far from the out-of-domain model, 
compared to a sensitivity analysis metric, 
suggesting that the out-of-domain model can provide a good generic initialization for the new domain.

\end{abstract}

\section{Introduction}

Neural Machine Translation (NMT) has supplanted Phrase-Based Machine Translation (PBMT) as the standard for high-resource machine translation. 
This has necessitated new domain adaptation methods,
because PBMT adaptation methods primarily rely 
on adapting the language model and phrase table
using interpolation or back-off schemes 
(see \autoref{related_work}).
Continued training \cite{Luong-Manning:iwslt15,FreitagA16}, 
also referred to as fine-tuning,
is one of the most popular methods for NMT 
adaptation, due to its strong performance.

In contrast to the PBMT literature,
little research has focused on why continued training is effective or on what happens to NMT models during continued training.
Motivated by domain adaptation analysis in PBMT \cite{haddow2012analysing, duh2010analysis,irvine2013measuring},
this work proposes a simple \emph{freezing subnetworks} technique 
and uses it to gain insight 
into how the various components of an NMT system
behave during continued training.

\begin{figure}[t]
	\centering
    \hspace{-10mm}    
	\includegraphics[width=1.0\linewidth]{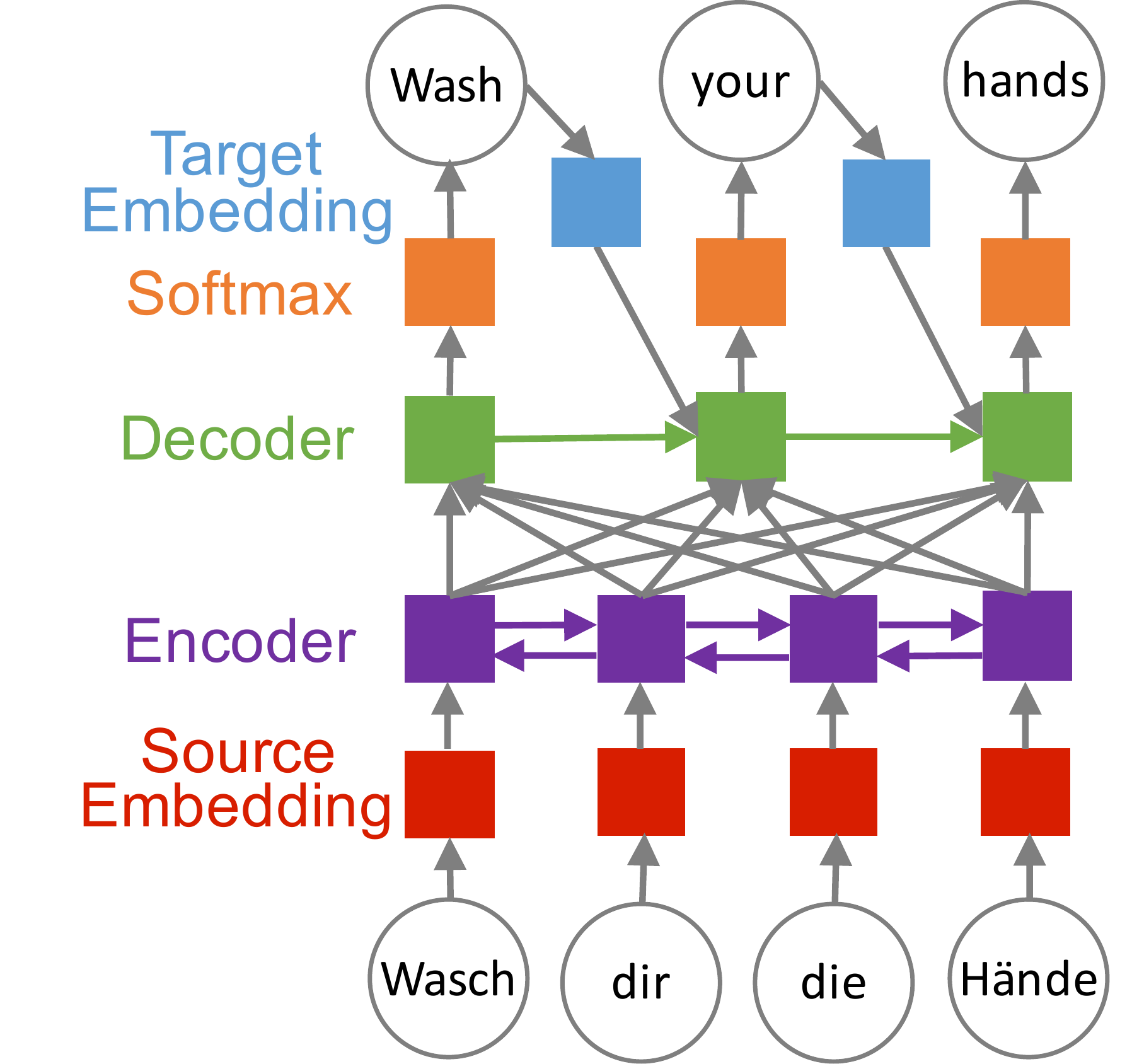}
    \caption{Visualization of an NMT system segmented into components.}
    \label{fig:nmt_diagram}
\end{figure}

\begin{table}[t]
	\begin{center}
\begin{tabular}{l r}
\toprule
Component    & {Size}  \\
\midrule
Target Embedding & 15.1M  \\ 
Softmax	     & 15.1M  \\ 
Decoder	     & 6.8M  \\ 
Encoder	     & 3.7M  \\ 
Source Embedding & 15.4M  \\
\midrule
Total & 56.0M\\
\bottomrule
\end{tabular}
\end{center}
    \caption{Number of parameters in each component.}
    \label{num_params_per_component}
\end{table}

We segment the model into five subnetworks, 
which we refer to as \emph{components}, 
denoted in \autoref{fig:nmt_diagram}: 
the source embeddings, encoder, decoder (which includes the attention mechanism), 
the softmax (used to denote the decoder output embeddings and biases), and the target embeddings. 

We freeze components one at a time during continued training to see how much the adaptation depends on each component.
We also experiment with freezing
everything except one component to 
determine each component's capacity to 
adapt to the new domain on its own.

In order to further analyze continued training, we examine the magnitude of change in model components during continued training of the network, 
under both normal and freezing training conditions. 
We also conduct sensitivity analysis of each component
to assist in interpreting these magnitudes.

Our NMT adaptation experiments are performed across three languages:
we translate from German, Korean, and Russian into English. Our out-of-domain models are trained on WMT and/or subtitles corpora, and we adapt each model to translate patent abstracts. 

\vspace{2mm}

\section{Related Work}\label{related_work}
\vspace{1mm}

Continued training has recently become a standard 
for domain or cross-lingual adaptation in several neural NLP applications. 
In PBMT, 
the most prominent methods focus on adapting the language model component 
\cite{moore2010intelligent}, and/or the translation model 
\cite{matsoukas2009discriminative,mansour2014translation,axelrod2011domain}, 
or on interpolating in-domain and out-of-domain models 
\cite{lu2007improving,foster2010discriminative,koehn2007experiments}. 

In contrast, the methods employed in NMT tend to utilize  continued training, 
which involves initializing the model with pre-trained weights 
(trained on out-of-domain data) and training/adapting it to the in-domain data. 
Among others,  \citet{Luong-Manning:iwslt15} and \citet{FreitagA16}
applied this method for domain adaptation. 
\citet{ChuDK17} mix in-domain and out-of-domain data during continued training in order to adapt to multiple domains.
Continued training has also been applied to cross-lingual transfer learning for NMT, with \citet{zoph-EtAl:2016:EMNLP2016} and \citet{nguyen-chiang:2017:I17-2} 
using it for transfer between high- and low-resource language pairs.

Continued training is effective on a range of data sizes.
In-domain gains have been shown with as few as dozens of in-domain training sentences \cite{micelibarone-EtAl:2017:EMNLP2017}, and recent work has explored continued training on single sentences \cite{farajian-EtAl:2017:WMT,kothur-knowles-koehn:2018:WNMT2018}.
 
Similar adaptation techniques are also employed in the field of Automatic Speech Recognition, 
where continued training has been the basis of cross-lingual transfer learning approaches 
\cite{grezl2014adaptation,kunze2017transfer}. 
Usually, the lower layers of the network, which perform acoustic modeling, are frozen and only the upper layers are updated. 
In a similar vein, other works \cite{swietojanski2014learning, N18-2080} 
adapt a network to a new domain by learning
additional weights that re-scale the hidden units.

\vspace{2mm}
\section{Data} \label{sec:data}
Our experiments are carried out across three language pairs, from Russian, Korean, and German into English. 
Basic statistics on the datasets used for our experiments are summarized in \autoref{tbl:data_sizes}. 
The three languages represent three different domain adaptation scenarios:
\begin{itemize}
 \item In German, both the in- and out-of-domain datasets are large. 
 \item In Russian, the in-domain dataset is large but the out-of-domain dataset is small. 
 \item In Korean, both in- and out-of-domain datasets are small. 
\end{itemize}

\begin{table}[t]
\centering
\begin{tabular}{l rrr}
\toprule
& & \multicolumn{2}{c}{Tokens} \\ \cmidrule(l){3-4} 
Dataset & \hspace{-2.5em}Sentences & {Source} & {Target} \\
\midrule
\multicolumn{4}{c}{Out-of-domain training sets} \\
\toEn{Ru} WMT       & \SI{25.2}{\mega\nothing} & \SI{563.9}{\mega\nothing}& \SI{595.9}{\mega\nothing} \\ 
\toEn{Ru} Subtitles	& \SI{25.9}{\mega\nothing} & \SI{179.8}{\mega\nothing} & \SI{212.4}{\mega\nothing} \\ 
\toEn{De} WMT	    & \SI{5.8}{\mega\nothing} & \SI{138.6}{\mega\nothing} & \SI{131.8}{\mega\nothing} \\    
\toEn{De} Subtitles & \SI{22.5}{\mega\nothing} & \SI{171.6}{\mega\nothing} & \SI{185.8}{\mega\nothing} \\
\toEn{Ko} Subtitles & \SI{1.4}{\mega\nothing} & \SI{11.5}{\mega\nothing} & \SI{11.9}{\mega\nothing} \\ 
\midrule
\multicolumn{4}{c}{In-domain training sets}\\
\toEn{Ru} WIPO   & \SI{29}{\kilo\nothing}  & \SI{620}{\kilo\nothing} & \SI{812}{\kilo\nothing} \\ 
\toEn{De} WIPO	 & \SI{821}{\kilo\nothing} & \SI{19}{\mega\nothing} & \SI{23}{\mega\nothing} \\ 
\toEn{Ko} WIPO   & \SI{81}{\kilo\nothing}  & \SI{2.2}{\mega\nothing} & \SI{2.0}{\mega\nothing} \\
\midrule
\multicolumn{4}{c}{In-domain test sets}\\
\toEn{Ru} WIPO   & \SI{3}{\kilo\nothing} & \SI{82}{\kilo\nothing} & \SI{109}{\kilo\nothing} \\ 
\toEn{De} WIPO	 & \SI{3}{\kilo\nothing} & \SI{132}{\kilo\nothing} & \SI{162}{\kilo\nothing} \\ 
\toEn{Ko} WIPO   & \SI{3}{\kilo\nothing} & \SI{187}{\kilo\nothing} & \SI{165}{\kilo\nothing} \\
\bottomrule
\end{tabular}
    \caption{Dataset statistics. The number of tokens is computed before segmentation into subwords. The in-domain development sets (not shown) have similar statistics to the test sets.}
    \label{tbl:data_sizes}
\end{table}

\begin{table*}[t]
\begin{center}
\begin{tabularx}{\textwidth}{l X}
\toprule
OpenSubtitles &  You're gonna need a bigger boat.\\ \midrule
WMT &  Intensified communication and sharing of information between the project partners enables the transfer of expertise in rural tourism.\\ \midrule
WIPO &  The films coated therewith, in particular polycarbonate films coated therewith, have improved properties with regard to scratch resistance, solvent resistance, and reduced oiling effect, said films thus being especially suitable for use in producing plastic parts in film insert molding methods. \\
\bottomrule
\end{tabularx}
\end{center}
    \caption{Example sentences to illustrate domain differences.}
    \label{tbl:examples}
\end{table*}

\subsection{Out-of-domain Data}

For our out-of-domain dataset we utilize the \texttt{OpenSubtitles2018} corpus \cite{TIEDEMANN16.62,LISON16.947}, which consists of translated movie subtitles.\footnote{\href{http://www.opensubtitles.org}{www.opensubtitles.org}} For \toEn{De} and \toEn{Ru}, we also use data from WMT 2017 \citep{wmt17},%
\footnote{\href{http://www.statmt.org/wmt17/}{statmt.org/wmt17}}
which contains data from several sources: Europarl (parliamentary proceedings) \citep{koehn2005europarl},%
\footnote{\href{http://www.statmt.org/europarl/}{statmt.org/europarl}}
News Commentary (political and economic news commentary),%
\footnote{\href{http://www.casmacat.eu/corpus/news-commentary.html}{casmacat.eu/corpus/news-commentary.html}}
Common Crawl (web-crawled parallel corpus), and the EU Press Releases.

We use the final \SI{2500}{} lines of \texttt{OpenSubtitles2018} for the development set. 
For German and Russian we also concatenate \texttt{newstest2016} as part of the development set.
\texttt{newstest2016}  consists of translated news articles released by WMT for its shared task.  In Korean, we rely only on the \texttt{OpenSubtitles2018} data.
See \autoref{tbl:examples} for example sentences from WMT and OpenSubtitles.  
\subsection{In-domain Data}

We perform adaptation into the World International Property Organization (WIPO) COPPA-V2 dataset \cite{junczys2016coppa}.\footnote{\href{http://www.wipo.int/patentscope/en/data/}{wipo.int/patentscope/en/data}} 
The WIPO data consist of parallel sentences from international patent application abstracts.
We reserve \SI{3000}{} lines each for the in-domain development and test sets. See \autoref{tbl:examples} for an example WIPO sentence.

\subsection{Data Preprocessing}

All our datasets were tokenized using the 
Moses\footnote{\href{http://www.statmt.org/moses/}{statmt.org/moses/}}
tokenizer.
Additionally, Korean text was segmented into words using the KoNLPy wrapper of the Mecab-Ko segmenter.\footnote{\href{http://www.konlpy.org/en/}{konlpy.org/en/}}

As a final preprocessing step, 
we train Byte Pair Encoding (BPE) segmentation models 
\citep{bpe} on the out-of-domain training corpus. 
We train separate BPE models for each language, 
each with a vocabulary size of \SI{30000}{}.
For each language, BPE is trained on the out-of-domain corpus only and then applied to the training, development, and test data for both out-of-domain and in-domain datasets.
This mimics the realistic setting where a generic, 
computationally-expensive-to-train NMT model is trained once. 
This NMT model is then adapted to new domains as they emerge, without retraining on the out-of-domain corpus.
Training BPE on the in-domain data would change the vocabulary and thus require re-building the model.

\section{Experimental Setup}

For all language pairs, 
we train systems on the out-of-domain data 
and select the best model parameters based on perplexity on the out-of-domain development set.
We then adapt the systems into our smaller, 
in-domain training sets.
We select the best model based on the WIPO
development set perplexity
and report results on the WIPO test sets.

\subsection{Continued Training}
We define continued training as: 
\begin{enumerate}
\vspace{-3mm}
\item Train a model until convergence on large out-of-domain bitext.
\vspace{-3mm}
\item Initialize a new model with the final parameters of Step 1.
\vspace{-3mm}
\item Train the model from Step 2 until convergence on in-domain bitext.
\end{enumerate}

\subsection{NMT Implementation and Settings}

Our neural machine translation systems are trained using \textsc{Sockeye} \citep{Sockeye:17}.%
\footnote{\href{https://github.com/awslabs/sockeye}{github.com/awslabs/sockeye}}
We use \textsc{Sockeye}'s built-in functionality for freezing parameters. 
We build RNN-based encoder--decoder models with attention \citep{bahdanau}, using a bidirectional RNN for the encoder. The encoder and decoder both have \SI{2}{} layers with LSTM hidden sizes of \SI{512}{}. Source and target word vectors are also of size \SI{512}{}. The number of parameters in each component are given in \autoref{num_params_per_component}.

While training the out-of-domain models, we apply dropout with \SI{10}{}$\%$ probability on the RNN layers. We apply label smoothing of \SI{.1}{}. We use \textsc{Adam} \cite{kingma14ADAM} as the optimizer, using a learning rate of \SI{0.0003}{} and a learning rate reduce factor of \SI{0.7}{}. We use a batch size of \SI{4096}{} words and create a checkpoint every \SI{4000}{} minibatches.

We do not use dropout or label smoothing during continued training because we do not want 
regularization to
bias our measurements of magnitude changes during continued training (see \autoref{E3}).
We note, however, that each would 
likely increase in-domain performance.
Our batch size during continued training is 128 sentences, and we create a checkpoint every half epoch. 
Our learning rate reduce factor for continued training is \SI{0.5}{}.
We run each continued training experiment over a set of learning rates ($0.1$, $0.01$, $0.001$, $0.0001$, $0.00001$)
and choose the best result based on the perplexity on the development set,
as previous work has suggested that
even when using \textsc{Adam}, continued training can be sensitive to learning rate \cite{farajian-EtAl:2017:WMT,LI18.195,kothur-knowles-koehn:2018:WNMT2018}.
We use dot product attention \cite{luong-pham-manning:2015:EMNLP}, which means we do not have a separate attention component; the attention is implicitly built into the decoder.

\section{Results and Analysis}\label{sec:analysis}

\subsection{Freezing One Component at a Time}\label{E1}
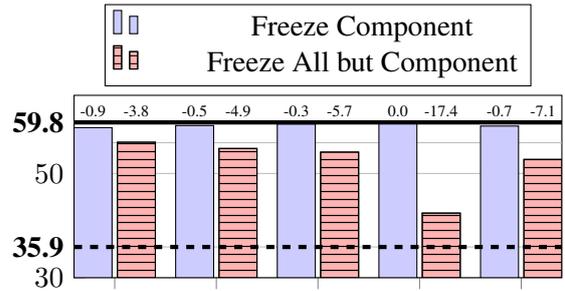
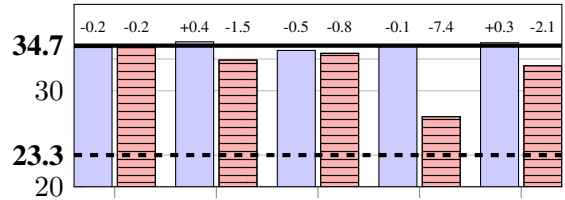
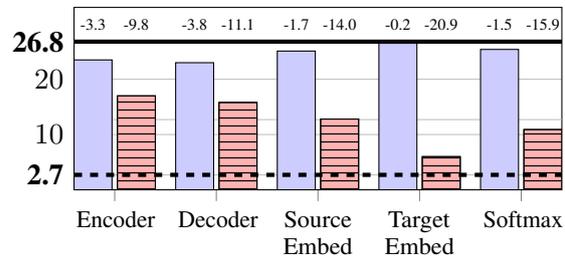
\begin{figure}[t]

\pgfplotstableread[row sep=\\,col sep=&]{
pos & model & freezeone & freezeall & diffone & difftwo \\
2 & Encoder & 58.8 & 56.0 & -0.9 & -3.8 \\
3 & Decoder & 59.2 & 54.8 & -0.5 & -4.9 \\
4 & Source Embed & 59.5 & 54.1 & -0.3 & -5.7 \\
5 & Target Embed & 59.8 & 42.4 & 0.0 & -17.4 \\
6 & Softmax & 59.1 & 52.7 & -0.7 & -7.1 \\
}\deendata

\pgfplotstableread[row sep=\\,col sep=&]{
pos & model & freezeone & freezeall & diffone & difftwo \\
2 & Encoder & 23.5 & 17.0 & -3.3 & -9.8 \\
3 & Decoder & 23.0 & 15.8 & -3.8 & -11.1 \\
4 & Source Embed & 25.1 & 12.8 & -1.7 & -14.0 \\
5 & Target Embed & 26.6 & 6.0 & -0.2 & -20.9 \\
6 & Softmax & 25.4 & 10.9 & -1.5 & -15.9 \\
}\koendata

\pgfplotstableread[row sep=\\,col sep=&]{
pos & model & freezeone & freezeall & diffone & difftwo \\
2 &Encoder& 34.5 & 34.5 & -0.2 & -0.2 \\
3 & Decoder & 35.1 & 33.2 & +0.4 & -1.5 \\
4 & {Source\\Embed} & 34.2 & 33.9 & -0.5 & -0.8 \\
5 & Target Embed & 34.6 & 27.3 & -0.1 & -7.4 \\
6 & Softmax & 35.0 & 32.6 & +0.3 & -2.1 \\
}\ruendata

	\centering
    \begin{subfigure}{\linewidth}


\begin{tikzpicture}
    \begin{axis}[
            ybar,
            bar width=.5cm,
            width=8cm,
            height=4cm,
            ymajorgrids=true,
            yminorgrids=true,
            legend style={at={(0.5,1.5)},
                anchor=north,legend columns=1,font=small},
            xticklabels={},
            tick pos=left,
            nodes near coords,
            nodes near coords align={vertical},
            visualization depends on=y \as \rawy,
            nodes near coords style=	{font=\tiny,shift={(axis direction cs:0,-\rawy+59)}},
            ymin=30,ymax=65,
            ytick={30,50},
            xlabel={},
            extra y ticks={35.9,59.8},
        	extra y tick labels={\textbf{35.9}, \textbf{59.8}},
        	extra y tick style={grid=minor},
            point meta=explicit symbolic,
        ]
        \addplot [style={black,fill=blue!20,mark=none}] table[x=pos,y=freezeone,meta=diffone]{\deendata};
        \addplot [style={black,postaction={pattern=horizontal lines},fill=red!30,mark=none}] table[x=pos,y=freezeall,meta=difftwo]{\deendata};
                
        \coordinate (A) at (axis cs:2,35.9);
		\coordinate (B) at (axis cs:0.1,59.8);
		\coordinate (O1) at (rel axis cs:0,0);
		\coordinate (O2) at (rel axis cs:1,0);
        \draw [black,sharp plot,dashed,ultra thick] (A -| O1) -- (A -| O2);
		\draw [black,sharp plot,ultra thick] (B -| O1) -- (B -| O2);
        
        \legend{Freeze Component, Freeze All but Component}
    \end{axis}
\end{tikzpicture}
        \caption{Results on WIPO \toEn{De}\vspace{1em}}
        \label{fig:freeze_de-en}
    \end{subfigure}
    ~ 
    \begin{subfigure}{\linewidth}

\begin{tikzpicture}
    \begin{axis}[
            ybar,
            bar width=.5cm,
            width=8cm,
            height=4cm,
            ymajorgrids=true,
            yminorgrids=true,
            legend style={at={(0.5,1.2)},
                anchor=north,legend columns=1,font=small},
            xtick=data,
            xticklabels={},
			x tick label style={font=\small,text width=1cm,align=center},
            tick pos=left,
            nodes near coords,
            nodes near coords align={vertical},
            visualization depends on=y \as \rawy,
            nodes near coords style=	{font=\tiny,shift={(axis direction cs:0,-\rawy+35)}},
            ymin=20,ymax=39,
            ytick={20,30},
            xlabel={},
            extra y ticks={23.3,34.7},
        	extra y tick labels={\textbf{23.3}, \textbf{34.7}},
            extra y tick style={grid=minor},
            point meta=explicit symbolic,
        ]
        \addplot [style={black,fill=blue!20,mark=none}] table[x=pos,y=freezeone,meta=diffone]{\ruendata};
        \addplot [style={black,postaction={pattern=horizontal lines},fill=red!30,mark=none}] table[x=pos,y=freezeall,meta=difftwo]{\ruendata};
        
        \coordinate (A) at (axis cs:2,23.3);
		\coordinate (B) at (axis cs:0.1,34.7);
		\coordinate (O1) at (rel axis cs:0,0);
		\coordinate (O2) at (rel axis cs:1,0);
        \draw [black,sharp plot,dashed,ultra thick] (A -| O1) -- (A -| O2);
		\draw [black,sharp plot,ultra thick] (B -| O1) -- (B -| O2);

    \end{axis}
\end{tikzpicture}
        \caption{Results on WIPO \toEn{Ru}\vspace{1em}}
        \label{fig:freeze_ru-en}
    \end{subfigure}
    ~ 
    \begin{subfigure}{\linewidth}


\begin{tikzpicture}
    \begin{axis}[
            ybar,
            bar width=.5cm,
            width=8cm,
            height=4cm,
            ymajorgrids=true,
            yminorgrids=true,
            legend style={at={(0.5,1.2)},
                anchor=north,legend columns=1,font=small},
            xtick=data,
			x tick label style={font=\small,text width=1cm,align=center},
            xticklabels from table= \koendata{model},
            tick pos=left,
            nodes near coords,
            nodes near coords align={vertical},
            visualization depends on=y \as \rawy,
            nodes near coords style=	{font=\tiny,shift={(axis direction cs:0,-\rawy+27)}},
            ymin=0,ymax=33,
            ytick={10,20},
            xlabel={},
            extra y ticks={2.7,26.8},
        	extra y tick labels={\textbf{2.7}, \textbf{26.8}},
        	extra y tick style={grid=minor},
            point meta=explicit symbolic,
        ]
        \addplot [style={black,fill=blue!20,mark=none}] table[x=pos,y=freezeone,meta=diffone]{\koendata};
        \addplot [style={black,postaction={pattern=horizontal lines},fill=red!30,mark=none}] table[x=pos,y=freezeall,meta=difftwo]{\koendata};
        
        \coordinate (A) at (axis cs:2,2.7);
		\coordinate (B) at (axis cs:0.1,26.8);
		\coordinate (O1) at (rel axis cs:0,0);
		\coordinate (O2) at (rel axis cs:1,0);
        \draw [black,sharp plot,dashed,ultra thick] (A -| O1) -- (A -| O2);
		\draw [black,sharp plot,ultra thick] (B -| O1) -- (B -| O2);
    \end{axis}
\end{tikzpicture}
        \caption{Results on WIPO \toEn{Ko}}
        \label{fig:freeze_ko-en}
    \end{subfigure}

\caption{BLEU scores 
when freezing only the denoted component (left solid bars) and when freezing all but the denoted component (right striped bars). The horizontal lines denote baselines: no adaptation (dashed) and full continued training (solid). The labels on top of each bar denote the difference from the full continued training baseline.}
\label{fig:freeze}
\end{figure}

Our first set of experiments measure the extent to which performance
depends on updating any given component in the model.
We perform continued training while freezing a single component (i.e. keeping that component fixed to the values from the out-of-domain model
used to initialize training
while adapting the rest of the components).
The results for this setting are shown in the solid left bars of \autoref{fig:freeze}.

For \toEn{De} and \toEn{Ru}, the out-of-domain models have reasonable performance on the in-domain test set. In these language pairs, freezing any single component has little impact on in-domain BLEU.
The worst change is \SI{-0.9}{} BLEU---when freezing the \toEn{De} encoder---and in some cases we see small gains of up to \SI{+0.4}{} BLEU. We interpret these gains as trivial (and possibly the result of variance) 
but there may be an NMT continued training scenario in which freezing could increase performance by acting as a regularizer  \citep[see][]{8268947}.

In \toEn{Ko}, where the out-of-domain model does poorly on the in-domain test set,
we see more substantial drops when freezing a component during continued training.
Freezing the decoder and encoder does the most harm (\SI{-3.8}{} and \SI{-3.3}{} BLEU, respectively), 
followed by the source embeddings and softmax components (\SI{-1.7}{} and \SI{-1.5}{} BLEU, respectively).

In all cases, freezing the target embeddings has very little impact (at most \SI{-0.2}{} BLEU, in \toEn{Ko}), 
suggesting that it is relatively unimportant during adaptation. These results show that the model and training procedure are very robust; continued training is able to find a local minimum for the new domain which has (nearly) equal performance to the one found in full training, even though an entire component is fixed to the initial out-of-domain model's values. 

This robustness suggests that caution is in order 
when attempting to interpret changes of any single component---in particular, changes in the surrounding components must also be considered.
For example, it appears that when the source embeddings are fixed, the encoder is able to compensate for the non-adapted source embeddings and adapt the system to interpret source tokens correctly in the new domain. Conversely, it appears that when the encoder is fixed, the source embeddings are able to adapt to produce vectors for source tokens which are interpreted correctly by the un-adapted encoder. Note that adaptation to source tokens in the new domain could theoretically occur in any un-frozen component, an idea further explored in the next section.

\subsection{Freezing All But One Component}\label{E2}
In our second set of experiments, we freeze all but one component during continued training
to see how much each component, in isolation, is able to
adapt the NMT system to the new domain.
The results are shown in \autoref{fig:freeze} (right striped bars).

We find that only adapting a single component is---somewhat surprisingly---not catastrophic in most cases. Adapting only the encoder, for example, still gives a gain of \SI{20.1}{} BLEU over the out-of-domain model (\SI{3.8}{} BLEU worse than full continued training) in German
and \SI{11.4}{} BLEU (\SI{0.2}{}~BLEU worse than full continued training) in Russian.

In \toEn{De} and \toEn{Ko}, we see that adapting just the encoder does the best, followed by the decoder, source embeddings, softmax, and target embeddings.
The trend in Russian is similar but with the decoder and source embeddings switched.

These experiments suggest the encoder is most able to adapt the model to a new domain in isolation. It is worth noting that the encoder achieves this despite being the component with the fewest parameters (3.7M). The target embeddings are least able to adapt the model to a new domain (consistent with \autoref{E1}). 

These experiments also show that the upper bound for adapting a single component is quite high, 
suggesting that the upper bound for adaptation techniques 
using monolingual data to adapt individual components 
could be quite high as well.
Of course, it seems unlikely that techniques using 
only monolingual data 
can achieve the same level of performance as when directly optimizing on bitext.

\begin{table}[t]
	\begin{center}
\begin{tabular}{l*{3}{S[table-format=1.4]}}
\toprule
             & {Russian}  &  {German} &  {Korean}  \\ 
\midrule
Softmax	     & 0.0347   &  0.0578 &   0.0650 \\ 
Encoder	     & 0.0236   &  0.0520 &   0.0654 \\ 
Decoder	     & 0.0209   &  0.0465 &   0.0594 \\ 
Source Embed & 0.0165   &  0.0417 &   0.0414 \\ 
Target Embed & 0.0141   &  0.0357 &   0.0422 \\ \bottomrule
\end{tabular}
\end{center}
    \caption{RMS change in each component when components are adapted jointly.}
    \label{distance_joint}
\end{table}

\begin{table}[t]
	\begin{center}
\begin{tabular}{l*{3}{S[table-format=1.4]}}
\toprule
             & {Russian}  &  {German} &  {Korean}  \\
\midrule
Softmax	     & 0.0345   &  0.2215 &   0.1031 \\
Encoder	     & 0.0516   &  0.2857 &   0.1494 \\
Decoder	     & 0.0419   &  0.2751 &   0.1122 \\
Source Embed & 0.0563   &  0.3045 &   0.0893   \\
Target Embed & 0.0714   &  0.2940 &   0.5777 \\
\bottomrule
\end{tabular}
\end{center}
    \caption{RMS change in each component when components are adapted individually.}
    \label{distance_independent}
\end{table}

\subsection{Magnitude of Changes During Continued Training}\label{E3}

We are interested in the overall magnitude of the changes experienced by each component during continued training, 
(i.e., how far each moves from the out-of-domain model)
and how those changes compare to the cases where only a single component was adapted.

We had two opposing hypotheses that could predict adaptation behavior
when only one component is being adapted (as in \autoref{E1}):
\begin{enumerate}
\vspace{-3mm}
  \item The portion of the network producing the component's input is fixed,
as is the portion of the network that 
interprets the component's output.
This suggests the component will
be somewhat constrained, in contrast to full continued training 
where the components may adapt jointly over time.
\vspace{-3mm}
  \item Since all other components are fixed, 
the adapting component has to bear all the responsibility 
for changing the entire model's behavior, 
requiring more drastic changes than it would have undergone
during full continued training.
\end{enumerate}

The root mean square (RMS) of the differences between each component in the initial out-of-domain model and the same component after continued training is shown in \autoref{distance_joint} (normal continued training) and \autoref{distance_independent} (trained individually).%
 While further work would be required to make any definitive statements, the results clearly favor the second hypothesis. 
The movement of individually adapted components tends to be larger than that of their counterparts in fully adapted models.

\subsection{Sensitivity Analysis}\label{E4}

To assist in interpreting the overall magnitude of changes experienced during continued training, 
we perform sensitivity analysis of each component of the initial, out-of-domain model.
In each experiment, zero-mean, independent Gaussian noise with fixed variance is added to every parameter in a single component
of the model. 
By varying noise levels, we show how much (random) movement is required 
to produce a given decrease in performance.\footnote{
\citet{bojar2010tackling} show that very low BLEU scores are not trustworthy.
Due to the very low BLEU score (\SI{2.7}{}) of the out-of-domain \toEn{Ko} system on the in-domain test set, we
use out-of-domain test sets for each language, where BLEU scores fall between \SI{11}{} and \SI{30}{}. 
This means that the BLEU scores for continued training (computed on the in-domain test set) 
are not directly comparable to the BLEU scores produced for sensitivity analysis. 
However, as the sensitivity analysis is used only as an aid in interpreting the general magnitude of BLEU shifts,
we view this as an acceptable compromise.}

\begin{figure}[t]
\centering
    \begin{subfigure}[b]{\linewidth}
        \includegraphics[width=.9\linewidth]{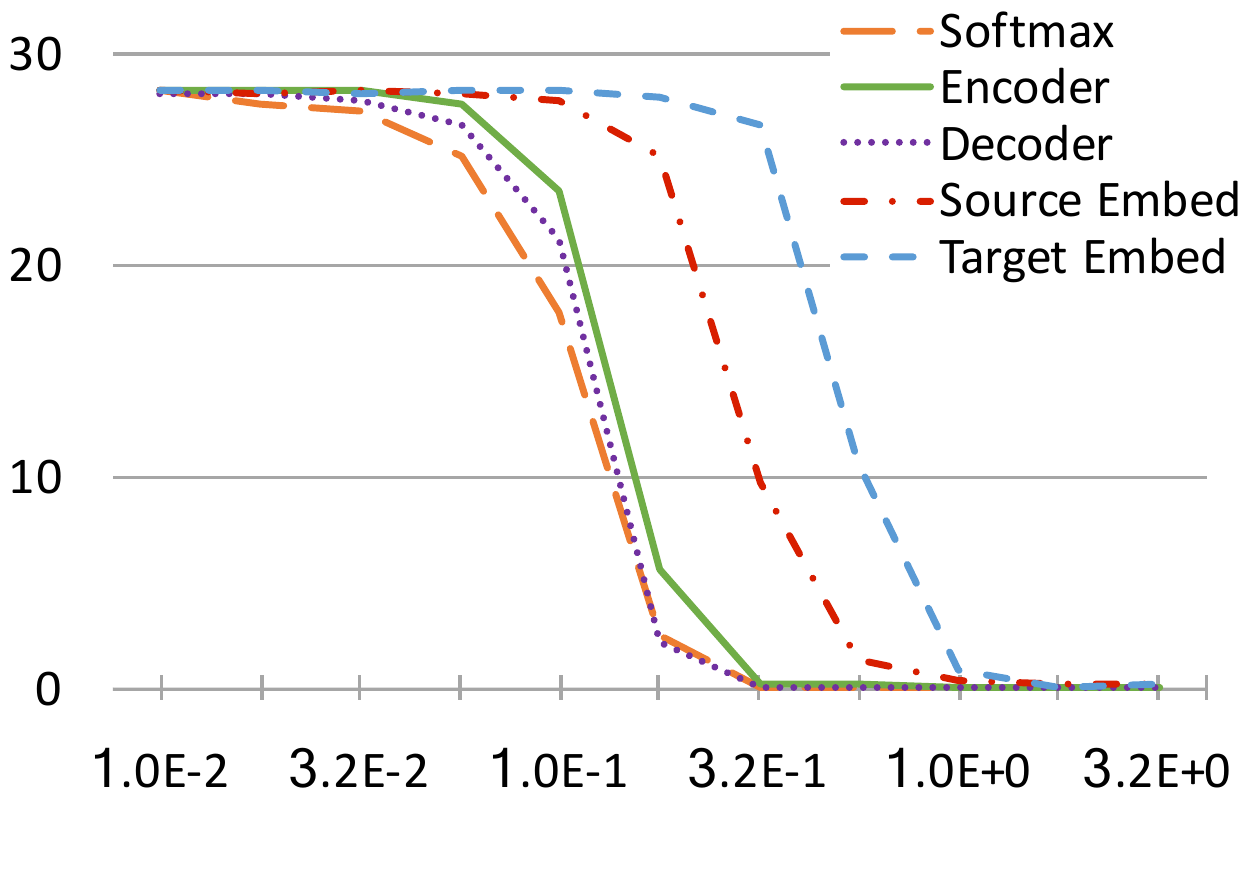}
       \vspace{-5mm}
        \caption{\toEn{Ru}}
        \label{fig:muck_ru_groups}
    \end{subfigure}
    ~ 
 
    \begin{subfigure}[b]{\linewidth}
        \includegraphics[width=.9\linewidth]{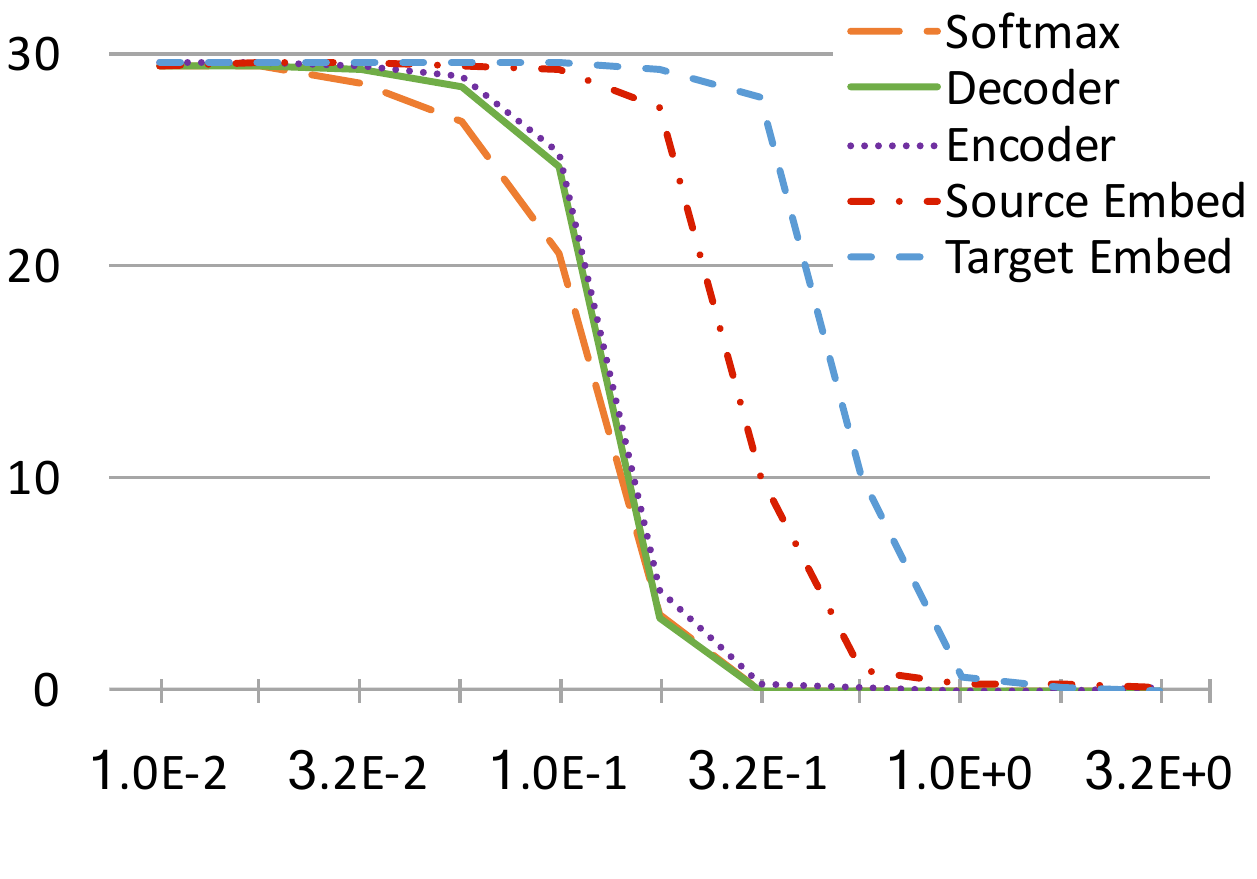}
         \vspace{-5mm}
        \caption{\toEn{De}}
        \label{fig:muck_de_groups}
    \end{subfigure}
    ~ 
    \begin{subfigure}[b]{\linewidth}
        \includegraphics[width=.9\linewidth]{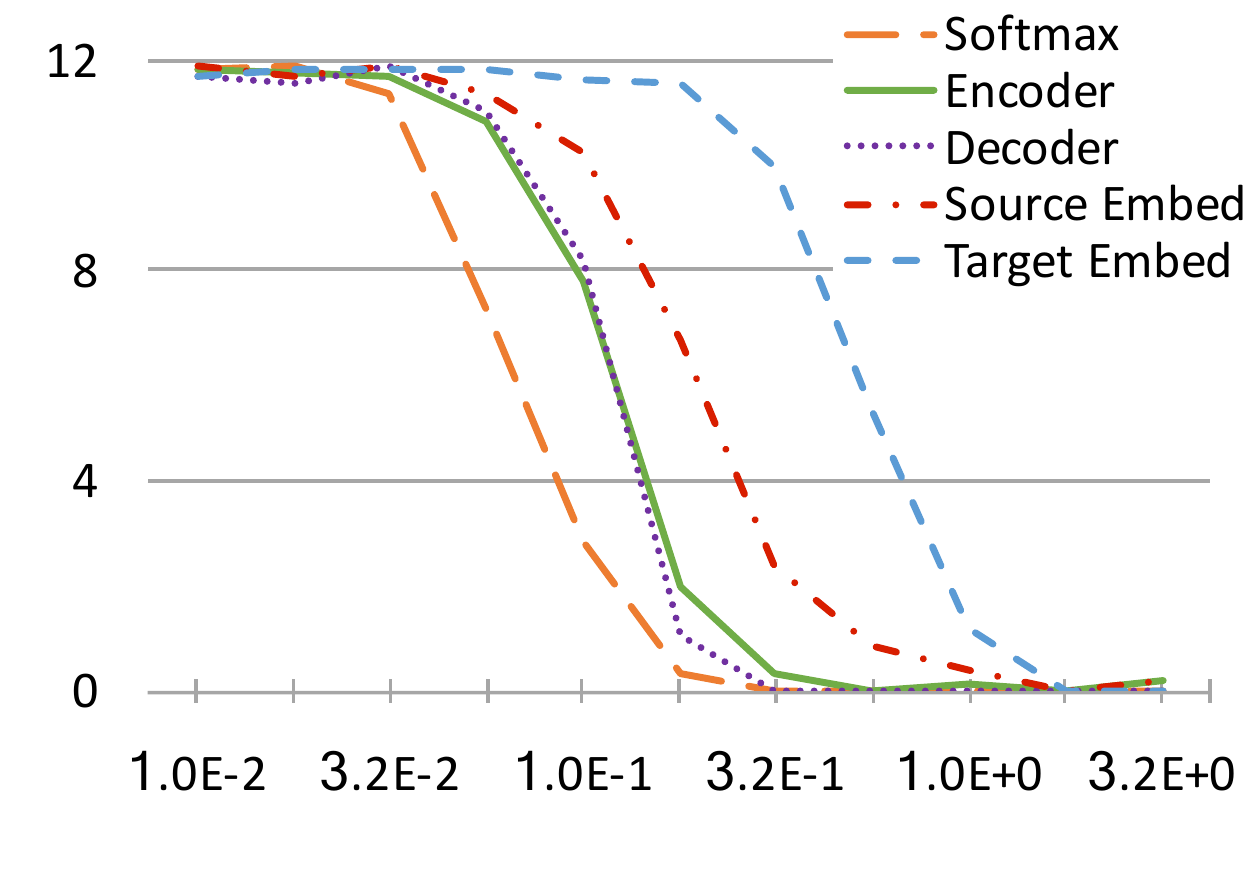}
            \vspace{-5mm}
        \caption{\toEn{Ko}}
        \label{fig:muck_ko_groups}
    \end{subfigure}

    \caption{Performance degradation (BLEU) as a function of noise (standard deviation) added to a given component.}
    \label{fig:muck_groups}
\end{figure}

\autoref{fig:muck_groups} 
shows the sensitivity plots for each component. 
\autoref{tab_E4} shows, for each component,
the (linearly interpolated) BLEU score decrease that would result 
from adding random noise of the same magnitude 
as the change observed in full continued training.

Considering the sensitivity of each component reveals several patterns. 
First, the 
most significant change in the network, compared to the sensitivity metric, 
is in the softmax component for all three languages. 
Second, these values are rather small compared to the overall improvements 
seen in continued training (\SI[retain-explicit-plus]{+23.0}{} in \toEn{De}, \SI[retain-explicit-plus]{+24.2}{} in \toEn{Ko}, and \SI[retain-explicit-plus]{+11.4}{} in \toEn{Ru}). 
This suggests that the in-domain model parameters are, on average, fairly close to the out-of-domain model used to initialize training; even though the out-of-domain model does not have a particularly high BLEU score, 
it is close to a good local minimum in the in-domain error surface.

\begin{table}[t]
	\begin{center}
\begin{tabular}{l*{3}{S[table-format=1.2]}}
\toprule
             & {Russian}  &  {German} &  {Korean}  \\
\midrule
Softmax	     & -1.29  &  -3.00 &   -5.49 \\
Encoder	     & -0.05  &  -0.78 &   -1.68 \\
Decoder	     & -0.23  &  -0.52 &   -1.05 \\
Source Embed & -0.12  &  -0.10 &   -0.22 \\
Target Embed & -0.08  &  -0.02 &   -0.04 \\
\bottomrule
\end{tabular}
\end{center}
    \caption{Sensitivity Analysis: Change in BLEU for random perturbation of magnitude corresponding to the distance each component moved during standard continued training.}
    \label{tab_E4}
\end{table}

\section{Conclusions}
 
This work presents and applies a simple \emph{freezing subnetworks} method to analyze continued training.

Freezing any single component during continued training has negligible effect on performance compared to full continued training. 
Furthermore, adapting only a \emph{single} component via continued training produces surprisingly strong performance in most cases, 
achieving most of the performance gain of full continued training. 
That is, continued training is able to adapt the overall system to 
a new domain by modifying only parameters in a single component. 
This finding goes against the intuitive hypothesis that source embeddings must account for domain changes in the source vocabulary, 
target embeddings must account for changes in the target vocabulary, etc. 

We note that the encoder and decoder, despite having the least parameters (3.7M and 6.8M, respectively, out of 56M), perform strongly across all  languages. This suggests further work on adapting only a subset of parameters may be warranted 
\citep[see also][]{N18-2080, michel2018extreme}.

We also perform sensitivity analysis of components and find that 
continued training does not move the model very far from the 
initial out-of-domain model, in the sense that random perturbations 
of the same magnitude cause only small performance drops 
on the out-of-domain test set. 
This suggests that the out-of-domain model, 
while not performing very well on the in-domain test set, 
is close to a good local minimum on the in-domain error surface.
This finding may explain the recent success of techniques which regularize a continued training model using the initial, out-of-domain model \cite{micelibarone-EtAl:2017:EMNLP2017,dakwale2017fine,W18-2705}.

\section*{Acknowledgements}
The authors would like to thank Lane Schwartz and Graham Neubig for their roles in organizing the  MT Marathon in the Americas (MTMA), where this work began. 
The authors would also like to thank Michael Denkowski and David Vilar for assistance with \textsc{Sockeye}.
This material is based upon work supported in
part by the DARPA LORELEI and IARPA MATERIAL programs. Brian Thompson is supported by the Department of Defense through the National Defense Science \& Engineering Graduate Fellowship (NDSEG) Program. Antonios Anastasopoulos is supported by the National Science Foundation (NSF) Award 1464553.
 
\thanksnostar{
Opinions, interpretations, conclusions and recommendations are those of the authors and are not necessarily endorsed by the United States Government.
Cleared for public release on 5 Sep 2018. Originator reference number RH-18-118777.
Case number 88ABW-2018-4431.
}

\bibliography{emnlp2018}
\bibliographystyle{acl_natbib_nourl}

\end{document}